\documentclass[letterpaper]{article} 
\usepackage{aaai2027}  
\usepackage[hyphens]{url}  
\usepackage{graphicx} 
\usepackage{natbib}  
\usepackage{caption} 
\usepackage{algorithm}
\usepackage{algorithmic}
\usepackage{amsmath}
\usepackage{amssymb}
\usepackage{multirow}
\usepackage{newfloat}
\usepackage{listings}
\DeclareCaptionStyle{ruled}{labelfont=normalfont,labelsep=colon,strut=off} 
\floatstyle{ruled}
\newfloat{listing}{tb}{lst}{}
\floatname{listing}{Listing}

\usepackage{booktabs}

\nocopyright
\title{Self-Improving Large Language Models via Progressive Experience Evolution}
\author{
    Shijie Ren\textsuperscript{\rm 1},
    Xiting Wang\textsuperscript{\rm 1}\corresponding,
    Meng Li\textsuperscript{\rm 1},
    Yujie Guo\textsuperscript{\rm 1},
    Yunhang Yao\textsuperscript{\rm 1},
    Ziheng Peng\textsuperscript{\rm 1}\\
    Xunlong Wang\textsuperscript{\rm 1},
    Yuetan Chen\textsuperscript{\rm 1},
    Haoyang Zhou\textsuperscript{\rm 1},
    Yunlong Liang\textsuperscript{\rm 2},
    Fandong Meng\textsuperscript{\rm 2}
}

\affiliations{
    \textsuperscript{\rm 1} Gaoling School of Artificial Intelligence, Renmin University of China \\
    \textsuperscript{\rm 2} Weixin AI, Tencent Inc, China\\
}

\begin{document}
 
\maketitle

\begin{abstract}
Large language models (LLMs) capable of self-improvement require not only effective policy optimization, but also a principled mechanism for transforming transient interaction experience into persistent model capabilities. Existing self-improvement paradigms remain fragmented: test-time methods can explicitly extract experience but cannot internalize it into model parameters, whereas training-time optimization methods can update model parameters but lack an explicit mechanism for accumulating transferable experience. Bridging these two paradigms requires a critical intermediate stage that remains underexplored, namely \emph{experience distillation}. To address this gap, we propose \textbf{SPEE} (\textbf{S}elf-\textbf{P}rogressive \textbf{E}xperience \textbf{E}volution), a unified post-training framework that sequentially performs explicit experience evolution followed by implicit policy optimization. During explicit experience evolution, SPEE reflects on trajectories collected from multiple interactions to extract, verify, and progressively evolve transferable experience, which is subsequently internalized into the policy through privilege-guided On-Policy Self-Distillation (OPSD). During implicit policy optimization, reward-driven reinforcement learning leverages these internalized priors to explore novel solution strategies. In the experience evolution stage, a continuously evolving global experience pool consolidates knowledge from both successful and failed trajectories, filters out low-utility experience, and mitigates post-hoc rationalization induced by individual trajectories. Experiments on five mathematical reasoning benchmarks demonstrate that SPEE consistently outperforms both test-time and training-time self-evolution baselines across three model scales. The source code is available at \url{https://github.com/rrrsj/SPEE}.
\end{abstract}

\section{Introduction}

Recent advances in LLMs have achieved remarkable progress across a wide range of tasks, including mathematical reasoning, code generation, and autonomous agents~\cite{achiam2023gpt,wei2022chain,yao2022react}. However, as model scales continue to grow, offline training pipelines have become increasingly insufficient, as they heavily rely on large amounts of high-quality data~\cite{kaplan2020scaling,ouyang2022training}. This limitation has motivated a shift toward self-evolving systems, where models are expected to autonomously adapt to new environments and acquire capabilities beyond their original training distributions~\cite{wang2023voyager,wang2026autonomous}. Despite rapid progress, how models achieve stable and continual self-evolution remains unclear. Existing approaches are mainly categorized by optimization objectives rather than the mechanisms underlying capability improvement~\cite{tao2survey}. We argue that the essence of self-evolution lies in transforming transient interaction signals into reusable and persistent model capabilities. In this work, we define \emph{experience} as compact and transferable knowledge abstracted from interaction trajectories, including reusable reasoning strategies, task-invariant constraints, and recurring failure patterns. Unlike raw trajectories, effective experience should remove instance-specific details, generalize across problems, and evolve with the policy. 

From the perspective of experience utilization, existing self-evolution paradigms can be broadly divided into test-time and training-time approaches. Test-time methods preserve demonstrations, reflections, or retrieved knowledge as additional context during inference~\cite{mohamed2025context,debnath2025comprehensive,wang2023selfconsistencyimproveschainthought}. Although effective, such experience remains external to model parameters and is constrained by context capacity, retrieval accuracy, and generalization ability~\cite{liu-etal-2024-lost,mueller-etal-2024-context}. In contrast, training-time methods, particularly reinforcement learning (RL), internalize rewarded behaviors through parameter updates and enable exploration of new behaviors~\cite{guo2025deepseek,yu2026dapo,shao2024deepseekmathpushinglimitsmathematical}. However, in reinforcement learning, the experience contained in successful and failed trajectories typically influences parameter updates only indirectly through sparse reward signals. Consequently, the model may require more extensive exploration to achieve better performance and can be highly sensitive to the quality of the initial model.

Together, these limitations point to a missing intermediate stage in self-evolution: Before reward-based reinforcement learning, textual experience should first be explicitly extracted from trajectories and internalized into the policy through dense supervision. This provides a stronger initialization for reinforcement learning, thereby improving exploration efficiency and potentially enabling a higher performance ceiling. Existing online policy self-distillation methods~\cite{hubotter2026reinforcement} provide a natural mechanism for such internalization by conditioning a teacher policy on instance-level trajectories and distilling the resulting enhanced behaviors into a student policy. However, these methods directly rely on complete reference trajectories rather than explicitly constructing transferable experience. Because such trajectories are tied to individual problems and often contain solution-specific details or answer information, the teacher may overly rely on instance-specific cues, increasing the risk of post-hoc rationalization rather than acquiring strategies that generalize across problems. Consequently, existing methods face two limitations: (1) complete trajectories entangle transferable knowledge with instance-specific content, making it difficult to form compact experience representations; and (2) they do not provide a complete pipeline for acquiring, refining, internalizing, and further optimizing experience, thereby limiting systematic experience accumulation across interactions.

To address this challenge, we propose \textbf{SPEE} (\textbf{S}elf-\textbf{P}rogressive \textbf{E}xperience \textbf{E}volution), a unified framework that transforms transient interaction experience into persistent model capabilities through progressive experience evolution and policy optimization. During explicit experience evolution, SPEE maintains a continuously evolving global experience pool that extracts and refines transferable knowledge from successful and failed trajectories, reducing post-hoc rationalization caused by instance-specific details. The evolved experience is then internalized into the policy through OPSD, enabling the accumulation of reusable knowledge without human intervention. During implicit policy optimization, reinforcement learning further improves the policy by encouraging exploration beyond existing experience. By coupling experience evolution with policy optimization, SPEE establishes the basis for a self-reinforcing loop: accumulated experience improves the policy, while the enhanced policy can generate richer experience for future rounds of evolution. Experiments on five mathematical benchmarks show that SPEE consistently outperforms both test-time and training-time baselines across three model scales, achieving up to 6.96\% improvement on qwen3-4b-base compared with the base model. These results demonstrate that SPEE provides an effective paradigm for continual self-improvement through progressive experience evolution and internalization.

\section{Method}
\label{sec:method}

\begin{figure*}[t]
\centering
\includegraphics[width=0.9\textwidth]{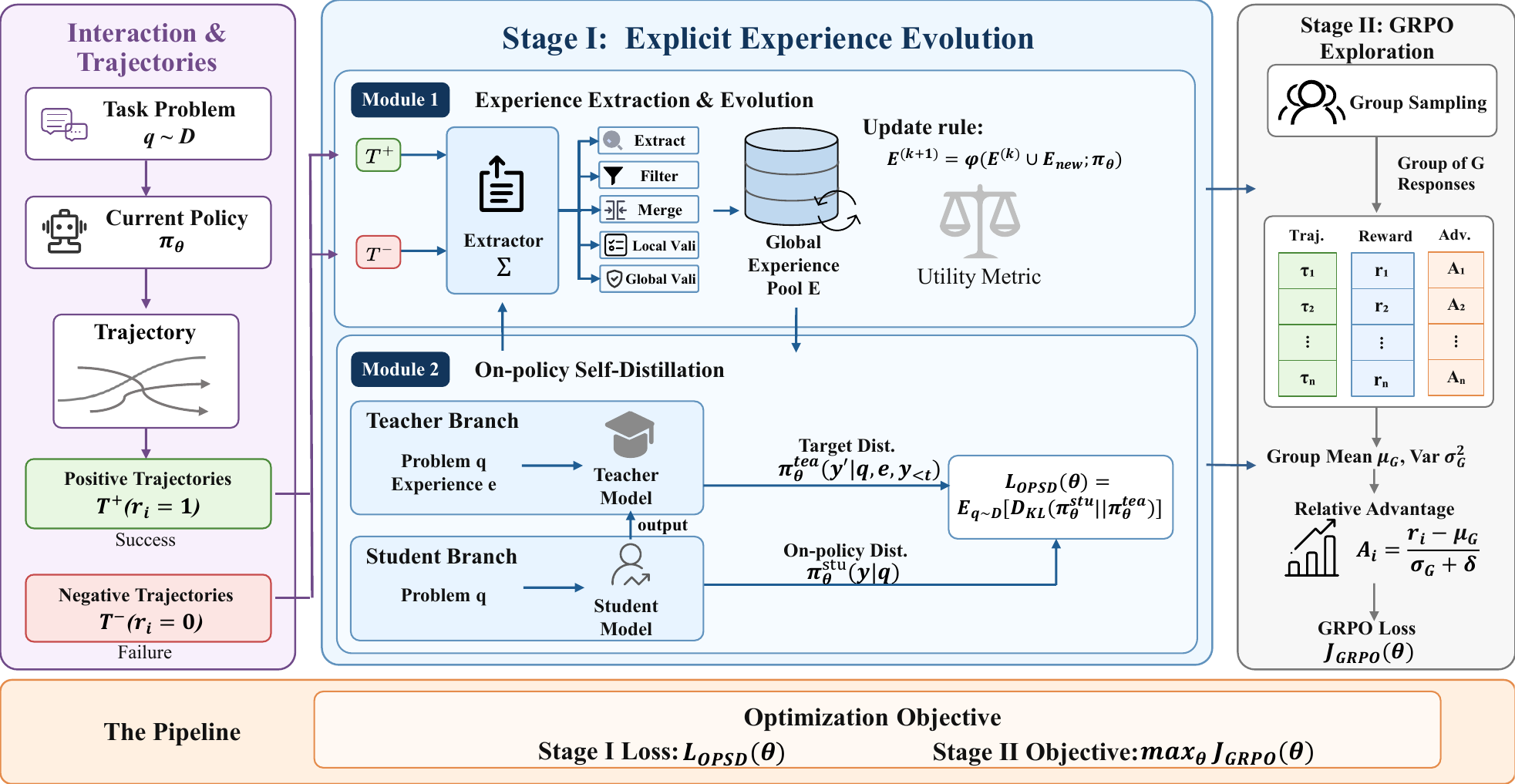}
\caption{Overview of SPEE. Interaction trajectories are used to evolve a global experience pool, whose distilled knowledge improves the policy and supports more effective reward-driven exploration in subsequent iterations.}
\label{fig:framework}
\end{figure*}

In this section, we first present an experience-centered perspective of LLMs' self-evolution, which serves as the conceptual foundation of SPEE. We then introduce explicit experience evolution, where transferable experience is extracted, evolved, and distilled into the policy. Finally, we present implicit policy optimization, where reward-driven exploration enables the policy to discover improved behaviors and generate richer experience for subsequent evolution, thereby closing the progressive self-improvement pipeline. The overall framework is illustrated in Figure~\ref{fig:framework}.

\subsection{Background: Experience-Centered Self-Evolution}
\label{sec:compression-view}

We revisit LLMs' self-evolution from the perspective of experience, viewing its objective as transforming transient interaction trajectories into persistent model capabilities. An effective self-evolution framework should not only extract experience more effectively, but also make better use of it.

Let $\mathcal{D}$ denote the input distribution, $\pi_\theta$ denote the policy parameterized by $\theta$. Given an input $q \sim \mathcal{D}$, the policy generates an output $y \sim \pi_\theta(\cdot \mid q)$, thereby forming an interaction trajectory $\mathbf{T}=\{\tau|\tau=(q,y)\}$. We further introduce $\mathbf{E}$ to represent the experience extracted from $\mathbf{T}$ by an experience extractor $\Sigma$, and $\mathbf{C}$ to represent the competence required to solve future tasks. To characterize the information relationships among these variables, we adopt mutual information as the measure. For two random variables $X$ and $Y$, $I(X;Y)$ represents how much information one variable provides about the other. Mutual information increases as the statistical dependence between the two variables becomes stronger and equals zero when they are independent. The extracted experience should preserve information that is predictive of competence on future tasks while discarding instance-specific details that do not generalize. This objective can be revisited as:
\begin{equation}
\min_{p_{\Sigma}(\mathbf{E}\mid\mathbf{T})}
I(\mathbf{T};\mathbf{E})
-
\lambda I(\mathbf{E};\mathbf{C}),
\label{eq:experience_ib}
\end{equation} 
where the first term encourages experience representations by penalizing unnecessary trajectory information. The second term preserves information predictive of future competence.

Under this view, different self-evolution methods can be interpreted as different approximations to the experience extractor $\Sigma$, differing primarily in how the extracted experience is represented and accumulated. Test-time methods explicitly preserve experience $e$ and provide as additional context during inference: $\pi_\theta(y\mid q,e)$. This enables utilization of experience, but the knowledge remains external to the model and is never internalized into parameters. Consequently, test-time methods improve experience utilization without transforming external experience into persistent model capabilities. Reinforcement learning instead attempts to compress interaction experience directly into model parameters through reward-driven optimization. Formally, the policy maximizes the expected trajectory reward: $J_{\mathrm{RL}}(\theta)
=
\mathbb{E}_{\tau\sim p_\theta}
\left[r(\tau)\right],
\label{eq:rl_objective}$ and is updated via gradient ascent:
\begin{equation}
\theta_{t+1}
=
\theta_t
+
\eta
\nabla_\theta J_{\mathrm{RL}}(\theta),
\label{eq:rl_update}
\end{equation} where $\eta$ denotes the learning rate. Rewarded behavioral patterns are gradually internalized into the policy. Standard reinforcement learning internalizes experience solely through reward signals, meaning that the information contained in trajectories can only be absorbed implicitly. When the sampled trajectories are of low quality, they may contain little information relevant to future competence: $I(\mathbf{T};\mathbf{C})\approx 0.$ As a result, the resulting supervision becomes weak, leading to sparse effective learning signals, inefficient exploration, and slow capability improvement.

The above analysis suggests that internalizing textual experience into the model before implicit policy optimization can not only improve sampling efficiency but also potentially lead to better performance. We refer to this process as \emph{experience distillation}, whose objective is:
\begin{equation}
\pi_{\theta^{'}}(y\mid q)
\leftarrow
\pi_\theta(y\mid q,e),
\label{eq:experience_distillation}
\end{equation}
thereby bridging explicit experience utilization and implicit policy optimization. Motivated by this perspective, we design SPEE to first explicitly evolve transferable experience and then progressively internalize it into the policy.

\subsection{Stage I: Explicit Experience Evolution}
\label{sec:opsd}

Rather than mechanically memorizing every detail, effective learners distill stable and generalizable principles from repeated attempts and reflection. Following this intuition, SPEE achieves the experience-distillation objective in Eq.~\ref{eq:experience_distillation} by maintaining a global experience pool rather than a static cache of trajectories or supervision. Specifically, SPEE first extracts transferable experience from trajectories and iteratively improves its reliability and transferability as the policy evolves. It then treats the refined experience as privileged information and internalizes it into the model parameters through OPSD. Accordingly, this stage consists of two components: experience evolution and experience distillation.

 \textbf{Experience evolution}. SPEE evolves experience over multiple rounds, at round \(k\), for each problem \(q\sim\mathcal{D}\), the current policy \(\pi_{\theta_k}\) samples \(G\) candidate responses:
\begin{equation}
y_i\sim\pi_{\theta_k}(\cdot\mid q), \quad i=1,\ldots,G.
\label{eq:candidate_sampling}
\end{equation}
Each response induces a trajectory \(\tau_i=(q,y_i)\) and receives a binary reward \(r_i=r(q,y_i)\). We partition the resulting trajectories into successful and failed sets:
\begin{equation}
\mathcal{T}^{+}(q) = \left\{\tau_i\mid r_i=1\right\}, \; \mathcal{T}^{-}(q) = \left\{\tau_i\mid r_i=0\right\}.
\label{eq:trajectory_partition}
\end{equation}
To accelerate the process, we randomly select only $S$ trajectories while preserving the empirical proportion of successful and failed responses.
The experience extractor $\Sigma_{\theta_k}$, implemented by the same model as the current policy under an experience-extraction prompt, converts each selected trajectory into a transferable experience item:
\begin{equation}
e_i^{(s_i)} = \Sigma_{\theta_k}(\tau_i,s_i), \; i\in\mathcal{I}_k(q).
\label{eq:experience_extraction}
\end{equation}
Positive items summarize effective reasoning patterns and reusable solution strategies, whereas negative items identify invalid reasoning patterns and recurrent failure modes. Retaining both is particularly important: successful trajectories reveal behaviors to reinforce, while failed trajectories help prevent repeated exploration of ineffective paths.

Since these items are extracted from the current policy’s behavior, they are treated as provisional hypotheses rather than final knowledge and may initially be incomplete, biased, or incorrect. We subsequently refine them through iterative evolution, allowing experience quality to improve alongside the policy. The newly extracted candidate set at round \(k\) is
\begin{equation}
\mathcal{E}_{\mathrm{new}}^{(k)} = \bigcup_{q\sim\mathcal{D}} \left\{ e_i^{(s_i)} \mid i\in\mathcal{I}_k(q) \right\}.
\label{eq:new_experience_set}
\end{equation}
We maintain the experience pool as a dynamic global memory. The newly extracted candidate set is integrated into a provisional pool:
\begin{equation}
\hat{\mathcal{E}}^{(k+1)} = \Phi\!\left( 
    \mathcal{E}^{(k)} \cup \mathcal{E}_{\mathrm{new}}^{(k)};\pi_{\theta_k}
\right),
\label{eq:provisional_evolve}
\end{equation}
where \(\Phi\) is an evolution operator implemented by the current model. It merges overlapping items, resolves redundant or conflicting content, and abstracts trajectory-specific observations into more transferable strategies.

To ensure experience quality, we measure each experience's marginal utility on a held-out probe set \(\mathcal{Q}_{\mathrm{pb}}\):
\begin{equation}
w^{(k)}(e) = \frac{1}{|\mathcal{Q}_{\mathrm{pb}}|} \sum_{q\in\mathcal{Q}_{\mathrm{pb}}}
\left[
    R_k\!\left(q,\hat{\mathcal{E}}_{e}^{(k+1)}\right) - R_k\!\left(q,\mathcal{E}^{(k)}\right)
\right],
\label{eq:utility}
\end{equation}
where $R_k(q,\mathcal{E}) = \mathbb{E}_{y\sim\pi_{\theta_k}(\cdot\mid q,\mathcal{E})} \big[r(q,y)\big]$
denotes the expected reward obtained by the current policy when conditioned on \(\mathcal{E}\). To balance effectiveness and efficiency, we apply a two-stage filter. Candidate items are first screened on a small subset of the probe set, and only those with positive utility are evaluated on the full probe set. Let
\begin{equation}
\widetilde{\mathcal{E}}^{(k)} =
\left\{
    e\in\hat{\mathcal{E}}_{\mathrm{new}}^{(k)} \mid w^{(k)}(e)>\epsilon
\right\}
\end{equation}
denote the accepted candidates, where \(\epsilon\geq 0\) is a validation threshold. The global pool is then updated as
\begin{equation}
\mathcal{E}^{(k+1)} = \Phi\!\left( \mathcal{E}^{(k)}\cup\widetilde{\mathcal{E}}^{(k)};
\pi_{\theta_k}
\right).
\label{eq:evolve}
\end{equation}
Repeated consolidation and validation yield the path
\begin{equation}
\mathcal{E}^{(0)} \rightarrow \mathcal{E}^{(1)} \rightarrow \cdots \rightarrow \mathcal{E}^{(M)},
\label{eq:experience_evolution_path}
\end{equation}
producing an increasingly reliable and transferable experience pool for subsequent distillation.

\paragraph{Privileged experience distillation.}

After several rounds of evolution, the pool $\mathcal{E}$ contains refined and high-utility experience. SPEE treats this experience as privileged information available only during training. We construct two branches with different information conditions. The teacher branch receives both the original problem and the global experience, while the student branch only observes the original problem. Both branches share the same underlying model parameters, while the teacher branch is detached from gradient computation. The student policy first generates trajectories following its current on-policy distribution:
\begin{equation}
y\sim\pi_{\theta}^{\mathrm{stu}}(\cdot|q).
\end{equation}
The sampled trajectories are then rescored at the token level by the experience-augmented teacher policy in an on-policy manner, thereby providing a stronger target distribution:
\begin{equation}
\pi_{\bar{\theta}}^{\mathrm{tea}}(y'|q)
\triangleq
\pi_{\bar{\theta}}(y'|q,e,y_{<t}).
\end{equation}
The student is optimized to improve its own on-policy behaviors by matching the teacher distribution. Specifically, SPEE minimizes the reverse KL divergence:
\begin{equation}
\begin{aligned}
\mathcal{L}_{\mathrm{OPSD}}(\theta)
&=
\mathbb{E}_{q\sim\mathcal D}
\left[
D_{\mathrm{KL}}
\left(
\pi_{\theta}^{\mathrm{stu}}(\cdot|q)
\Vert
\pi_{\bar{\theta}}^{\mathrm{tea}}(\cdot|q)
\right)
\right].
\end{aligned}
\end{equation}

\subsection{Stage II: Implicit Policy Optimization}
\label{sec:implicit-policy-optimization}

Experience distillation internalizes the evolved experience into the policy, providing a strong initialization for further self-improvement. However, because this process learns from experience already represented in the pool, its gains are inherently limited by the pool’s current coverage. To further expand the policy's capabilities, we apply Group Relative Policy Optimization (GRPO) to the distilled policy. Through reward-driven exploration, this stage discovers high-reward behaviors beyond the existing experience pool while directly improving the policy.

For each problem $q \sim \mathcal{D}$, we sample a group of $G$ responses from the behavior policy. Let $r_i = r(q,y_i)$ denote the reward assigned to response $y_i$. GRPO computes the mean and variance of the rewards within each group as
\begin{equation}
\mu_G
=
\frac{1}{G}
\sum_{j=1}^{G} r_j,
\qquad
\sigma_G^2
=
\frac{1}{G}
\sum_{j=1}^{G}
\left(r_j-\mu_G\right)^2.
\label{eq:grpo_statistics}
\end{equation}
The group-relative advantage associated with response $y_i$ is then defined as
\begin{equation}
A_i
=
\frac{r_i-\mu_G}{\sigma_G+\delta},
\label{eq:grpo_adv}
\end{equation}
where $\delta > 0$ is a small constant ensuring numerical stability.
For each generated token $y_{i,t}$, the importance-sampling ratio between the current policy and the behavior policy is
\begin{equation}
\rho_{i,t}(\theta)
=
\frac{
\pi_{\theta}
\left(
y_{i,t}
\mid
q,y_{i,<t}
\right)
}{
\pi_{\theta_{\mathrm{old}}}
\left(
y_{i,t}
\mid
q,y_{i,<t}
\right)
}.
\label{eq:grpo_ratio}
\end{equation}
Following the clipped policy optimization objective, the token-level surrogate term is:
\begin{equation}
\ell_{i,t}^{\mathrm{clip}}(\theta)
=
\min
\left\{
\rho_{i,t}(\theta) A_i,
\bar{\rho}_{i,t}(\theta) A_i
\right\},
\label{eq:grpo_clip}
\end{equation}
where the clipped importance ratio is given by $\bar{\rho}_{i,t}(\theta)
=
\operatorname{clip}
\left(
\rho_{i,t}(\theta),
1-\epsilon,
1+\epsilon
\right),$ and $\epsilon$ denotes the clipping coefficient. The overall GRPO objective is:
\begin{equation}
\mathcal{J}_{\mathrm{GRPO}}(\theta)
=
\mathbb{E}_{
\substack{
q \sim \mathcal{D},\\
\mathbf{y} \sim
\pi_{\theta_{\mathrm{old}}}(\cdot \mid q)
}
}
\left[
\frac{1}{G}
\sum_{i=1}^{G}
\frac{1}{|y_i|}
\sum_{t=1}^{|y_i|}
\ell_{i,t}^{\mathrm{clip}}(\theta)
\right],
\label{eq:grpo}
\end{equation}
The policy is optimized by maximizing $\mathcal{J}_{\mathrm{GRPO}}(\theta)$.

Unlike directly applying GRPO to the base checkpoint, SPEE initializes policy optimization from the policy obtained through self-evolving experience distillation, increases the likelihood of sampling rewarded trajectories and, more importantly, groups whose responses receive different reward values. Such groups produce non-zero relative advantages and therefore provide informative policy-gradient signals. In contrast, when all responses in a group receive the same reward, their normalized advantages become zero, resulting in an uninformative optimization step. By moving the policy toward high-reward regions before reinforcement learning, experience distillation improves exploration efficiency. We further analyze this effect in our experiments. It is worth noting that the second stage can also employ other variants of GRPO, which may yield better performance. For simplicity in both method design and experimental setup, we use only the standard GRPO algorithm in this work.

\section{Experiments}
\label{sec:experiments}

\subsection{Setting}
We evaluate SPEE on three scales of the Qwen3 family: \texttt{qwen3-1.7b/4b/8b-base}. We control the number of training steps and the total number of sampled rollouts across different methods. We train models on DAPO-math-17k~\cite{yu2026dapo} and report results on five widely used mathematical benchmarks: AIME24/25, GSM8K, MATH500, and Minerva Math~\cite{cobbe2021trainingverifierssolvemath,lightman2023lets,aime24,aime25,lewkowycz2022solving}. These benchmarks cover competition-level olympiad problems, grade-school word problems, and general mathematical reasoning tasks. For GSM8K, MATH500, and Minerva Math, we report pass@1 results, while for AIME 2024 and AIME 2025, we report pass@16 average results, along with the performance change $\Delta$ relative to the base model checkpoint. We compare SPEE with four representative baselines: (i) the Base model; (ii) Domain Prompt, a test-time baseline which applies a manually designed mathematical reasoning prompt during inference; (iii) GRPO~\cite{shao2024deepseekmathpushinglimitsmathematical}, a reinforcement learning baseline; and (iv) SDPO~\cite{hubotter2026reinforcement}, a self-distillation-based method. 

\subsection{Main Results}
Table~\ref{tab:main_results} presents the main results. Across three model scales, SPEE achieves the highest average accuracy. Compared with the base model checkpoints, SPEE yields improvements of +4.87, +6.96, and +6.53 percentage points on the 1.7B, 4B, and 8B models, respectively. Compared with GRPO, SPEE achieves superior performance at all model scales, with an average improvement of 1.16\%. Although GRPO can discover high-quality behaviors through reward-driven exploration, its effectiveness is constrained by the capability of the initial policy. In contrast, SPEE internalizes experience into the model parameters through experience distillation, thereby improving the efficiency of subsequent policy optimization. SPEE also consistently outperforms SDPO. Unlike directly distilling reference answers, which may cause the model to overfit specific reasoning paths and produce post-hoc rationalizations, SPEE employs a continuously evolving global experience pool to extract more abstract and transferable reasoning patterns. The comparison with Domain Prompt further highlights the importance of parameter-level experience accumulation. Overall, these results validate SPEE as an effective framework for supporting the continual self-evolution of large language models.

\begin{table*}[t]
    \centering
    \small
    \setlength{\tabcolsep}{8pt}
    \renewcommand{\arraystretch}{1.05}

    \begin{tabular}{llccccccc}
        \toprule
        \textbf{Model}
        & \textbf{Method}
        & \textbf{AIME24}
        & \textbf{AIME25}
        & \textbf{GSM8K}
        & \textbf{MATH500}
        & \textbf{MINA}
        & \textbf{Average}
        & \textbf{$\Delta$} \\
        \midrule

  \multirow{5}{*}{\texttt{qwen3-1.7b}}
        & Base
        & 1.45 & 4.16 & 71.44 & 39.90 & 11.49
        & 25.69 & -- \\

        & Domain Prompt
        & 1.67 & 2.50 & 74.07 & 39.30 & 13.20
        & 26.15 & +0.46 \\

        & GRPO
        & 5.00 & 6.67 & 80.81 & 42.50 & \textbf{14.07}
        & 29.81 & +4.12 \\

        & SDPO
        & 0.00 & 1.67 & 71.92 & 39.40 & 9.56
        & 24.51 & -1.18 \\

        & \textbf{SPEE (Ours)}
        & \textbf{5.21} & \textbf{7.71}
        & \textbf{82.13} & \textbf{43.80} & 13.97
        & \textbf{30.56} & \textbf{+4.87} \\

        \midrule

        \multirow{5}{*}{\texttt{qwen3-4b}}
        & Base
        & 7.91 & 6.46 & 73.10 & 45.53 & 15.54
        & 29.71 & -- \\

        & Domain Prompt
        & 7.29 & 9.17 & 85.59 & 48.00 & 14.06
        & 32.82 & +3.11 \\

        & GRPO
        & 10.00 & 4.79 & 89.73 & 49.75 & 19.49
        & 34.75 & +5.04 \\

        & SDPO
        & 6.67 & \textbf{10.83} & 83.57 & 48.65 & 18.75
        & 33.69 & +3.98 \\

        & \textbf{SPEE (Ours)}
        & \textbf{11.04} & 8.75
        & \textbf{90.83} & \textbf{52.80} & \textbf{19.94}
        & \textbf{36.67} & \textbf{+6.96} \\

        \midrule

        \multirow{5}{*}{\texttt{qwen3-8b}}
        & Base
        & 7.50 & 6.87 & 80.38 & 46.85 & 19.30
        & 32.18 & -- \\

        & Domain Prompt
        & 4.17 & 6.87 & 89.82 & 48.50 & 21.97
        & 34.27 & +2.09 \\

        & GRPO
        & 9.17 & 11.29 & 92.47 & 52.80 & 23.86
        & 37.92 & +5.74 \\

        & SDPO
        & 3.33 & 11.25 & 90.26 & 49.15 & 21.88
        & 35.17 & +2.99 \\

        & \textbf{SPEE (Ours)}
        & \textbf{10.60} & \textbf{12.08}
        & \textbf{93.14} & \textbf{53.70} & \textbf{24.06}
        & \textbf{38.71} & \textbf{+6.53} \\

        \bottomrule
    \end{tabular}
    \caption{Performance comparison of different post-training methods across mathematical benchmarks.}
    \label{tab:main_results}
\end{table*}

\begin{figure}[t]
\centering
\includegraphics[width=0.47\textwidth]{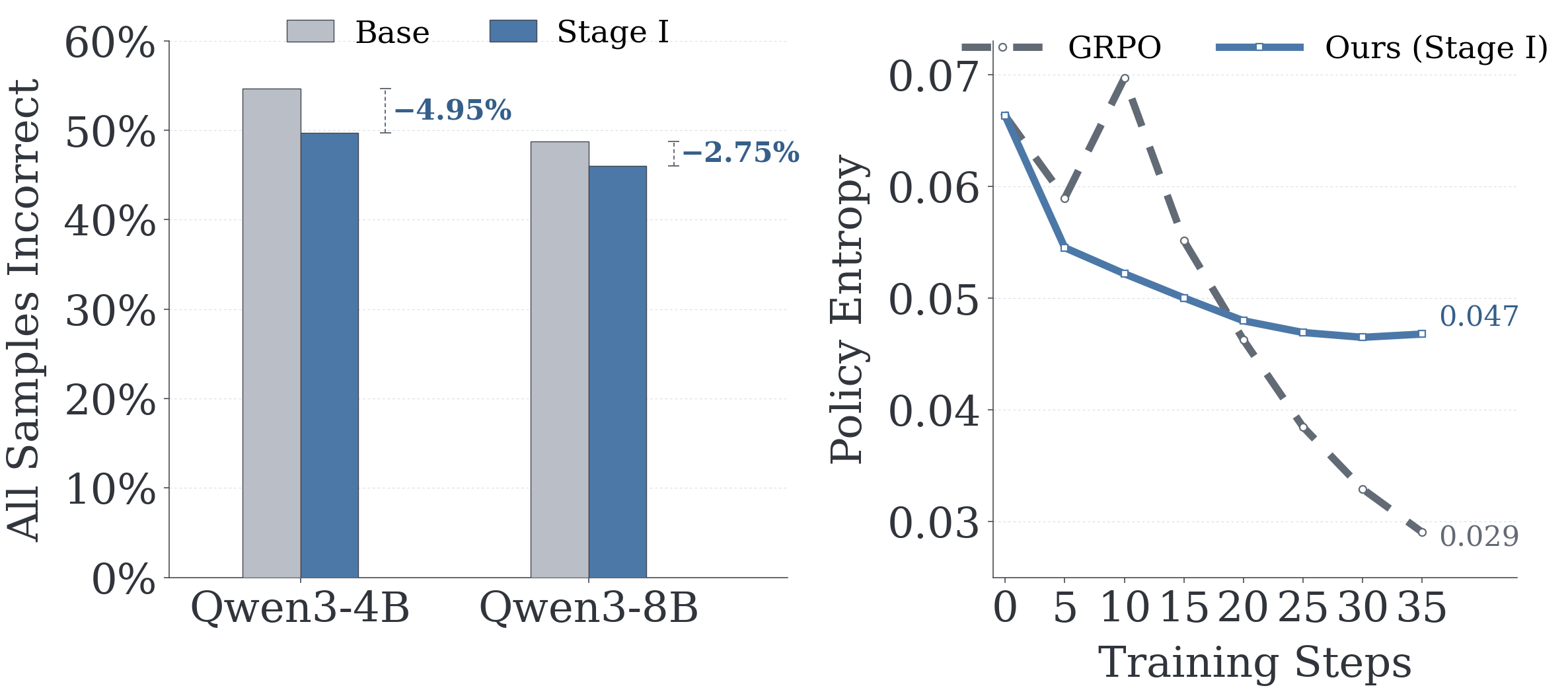}
\caption{(Left) Proportion of problems for which all eight sampled responses are incorrect before and after Stage I. Experience distillation reduces the frequency of all-incorrect response groups. (Right) Policy entropy throughout training.}
\label{fig:entropy}
\end{figure}

\begin{figure}[t]
\centering
\includegraphics[width=0.45\textwidth]{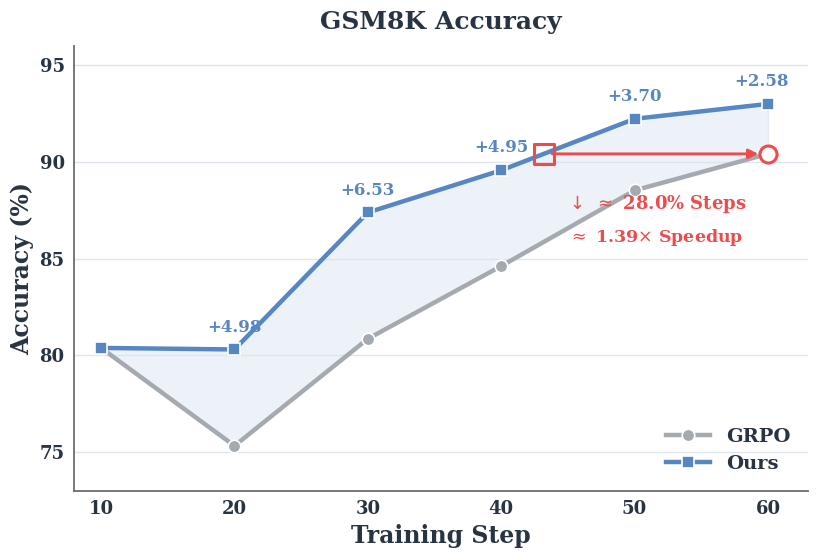}
\caption{Data efficiency comparison between SPEE and GRPO. SPEE achieves comparable or superior performance with substantially fewer training trajectories.}
\label{fig:data_efficiency}
\end{figure}

\subsection{Ablation Study}
To validate the effectiveness of each component in our method, we conduct an ablation study. Table~\ref{tab:ablation} evaluates the contributions of the two main components of SPEE. Removing Stage II decreases the average performance from 36.67\% to 35.16\% on the 4B model and from 38.71\% to 36.14\% on the 8B model, demonstrating the importance of reward-driven policy optimization. Removing the shared experience pool leads to a further performance drop, with the average accuracy decreasing to 34.13\% and 35.48\% on the 4B and 8B models, respectively. These results indicate that both components are essential to SPEE, while the shared experience pool plays a particularly important role in accumulating and transferring reusable experience across different problems. Overall, the complete SPEE framework achieves the best average performance at both model scales.

\begin{table*}[t]
    \centering
    \small
    \setlength{\tabcolsep}{6.5pt}
    \renewcommand{\arraystretch}{1.05}
    \begin{tabular}{llccccccc}
        \toprule
        \textbf{Model} & \textbf{Method} & \textbf{AIME24} & \textbf{AIME25} &
        \textbf{GSM8K} & \textbf{MATH500} & \textbf{MINA} &
        \textbf{Average} & \textbf{$\Delta$} \\
        \midrule
        \multirow{4}{*}{\texttt{qwen3-4b}}
        & Base
        & 7.91 & 6.46 & 73.10 & 45.53 & 15.54 & 29.71 & -- \\
        
        & w/o Stage2
        & 6.46 & \textbf{8.75} & 89.61 & 50.00 & \textbf{20.96}
        & 35.16 & +5.45 \\
        
        & w/o Shared Experience Pool
        & 6.67 & 8.33 & 87.32 & 49.74 & 18.57
        & 34.13 & +4.42 \\
        
        & \textbf{SPEE (Ours)}
        & \textbf{11.04} & \textbf{8.75} & \textbf{90.83} &
        \textbf{52.80} & 19.94 & \textbf{36.67} & \textbf{+6.96} \\
        \midrule
        \multirow{4}{*}{\texttt{qwen3-8b}}
        & Base
        & 7.50 & 6.87 & 80.38 & 46.85 & 19.30 & 32.18 & -- \\
        
        & w/o Stage2
        & 7.08 & 6.67 & 92.27 & 52.60 & 22.06
        & 36.14 & +3.96 \\
        
        & w/o Shared Experience Pool
        & 5.00 & 8.13 & 88.38 & 52.35 & 23.53
        & 35.48 & +3.30 \\
        
        & \textbf{SPEE (Ours)}
        & \textbf{10.60} & \textbf{12.08} & \textbf{93.14} &
        \textbf{53.70} & \textbf{24.06} & \textbf{38.71} &
        \textbf{+6.53} \\
        \bottomrule
    \end{tabular}
    \caption{Ablation study of different components in SPEE. 
    $\Delta$ denotes the improvement over the corresponding base model.}
    \label{tab:ablation}
\end{table*}

\subsection{Training dynamics}
\label{sec:efficiency}

\textbf{Rollout Diversity.} A key premise of our method is that Stage I improves policy performance without prematurely reducing policy entropy. To validate this premise, we compare the base policy with the policy obtained after Stage I in terms of both the proportion of sampled responses receiving positive rewards and the policy entropy during generation. We perform 8 rollouts for each problem. As shown in Figure~\ref{fig:entropy} (left), Stage I substantially reduces the proportion of problems for which all sampled responses are incorrect. Consequently, the policy is more likely to generate response groups containing both correct and incorrect solutions, thereby providing more informative gradient signals for GRPO. As shown in Figure~\ref{fig:entropy} (right), the policy after Stage I still maintains an entropy level comparable to that of the base model. These results indicate that experience distillation shifts the policy toward higher-reward regions while avoiding premature policy collapse or excessive restriction of response diversity.

\textbf{Data Efficiency.} As shown in Figure~\ref{fig:data_efficiency}, benefiting from the dense supervision provided in Stage I and the higher effective sampling efficiency achieved in Stage II, SPEE attains performance comparable to or even better than GRPO while using substantially fewer training trajectories. SPEE first extracts transferable knowledge from both successful and failed trajectories during the experience evolution stage and internalizes it into the policy in advance, enabling the initialized policy to generate more effective responses. Experimental results show that, at the same performance level, SPEE requires approximately 28\% less training data than GRPO, demonstrating superior data utilization efficiency.

\subsection{Effect of Experience Evolution Iterations}
\label{app}

\begin{figure}[t]
\centering
\includegraphics[width=0.47\textwidth]{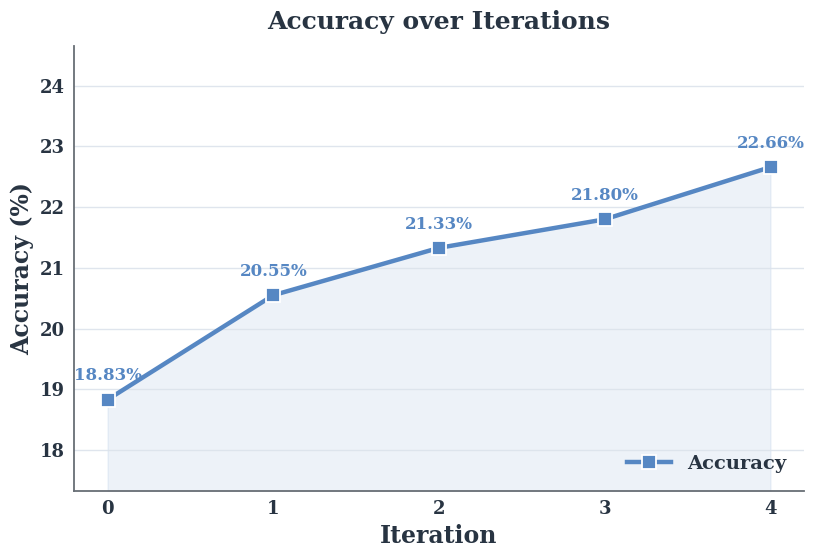}
\caption{Accuracy under different numbers of experience evolution iterations. Iteration 0 denotes the setting without experience evolution. Performance consistently improves as the experience pool undergoes additional rounds of evolution.}
\label{fig}
\end{figure}

To investigate how progressive experience evolution affects the sampling capability of the policy, we conduct an additional controlled experiment by varying the number of experience evolution iterations from 0 to 4 while keeping the remaining configuration unchanged. We report \emph{sampling accuracy}, defined as the proportion of sampled responses that are correct and receive a positive reward. Therefore, this metric reflects the ability of the current policy to generate correct trajectories during sampling, rather than its final evaluation accuracy on downstream benchmarks. Iteration 0 denotes sampling without experience evolution, while each subsequent iteration introduces an additional round of experience extraction, consolidation, filtering, and validation. As shown in Figure~\ref{fig}, the sampling accuracy consistently increases as the experience pool evolves. Specifically, it improves from 18.83\% at iteration 0 to 20.55\%, 21.33\%, 21.80\%, and 22.66\% after one to four evolution iterations, respectively. After four iterations, the sampling accuracy increases by 3.83 percentage points, corresponding to a relative improvement of approximately 20.34\% over the setting without experience evolution. These results indicate that repeated experience evolution progressively improves the policy's ability to sample correct trajectories. Rather than merely accumulating additional textual information, the evolving experience pool consolidates overlapping knowledge, filters unreliable items, and preserves transferable reasoning patterns that provide more effective guidance during generation. As the quality of the experience pool improves, the policy is more likely to produce positively rewarded responses, thereby providing higher-quality trajectories and more informative learning signals for subsequent policy optimization. Although the main experiments adopt a single evolution round for computational efficiency, the consistent improvement in sampling accuracy suggests that multiple rounds of experience evolution may further strengthen the self-improvement loop.

\subsection{Qualitative Analysis}
\begin{figure*}[t]
\centering
\includegraphics[width=0.86\textwidth]{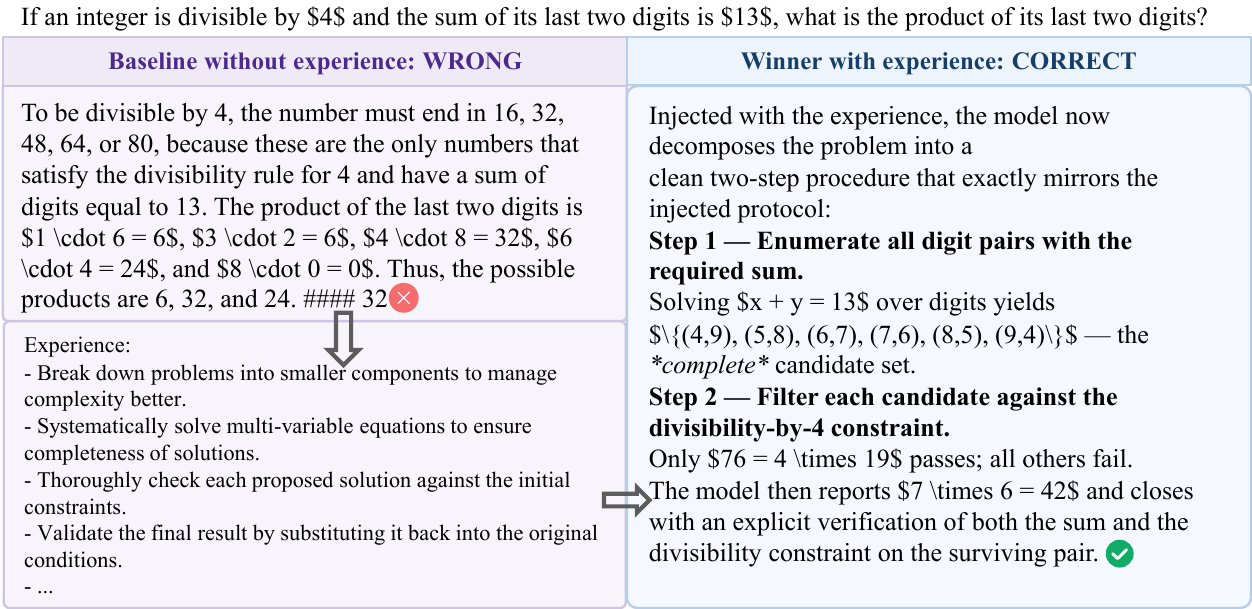}
\caption{A representative case illustrating the effect of experience-guided reasoning. Without experience, the model constructs an incomplete candidate set and produces an incorrect answer. After incorporating evolved experience, the model systematically enumerates all feasible digit pairs and verifies each candidate against the divisibility constraint, obtaining the correct answer.}
\label{fig:case_study}
\end{figure*}

Figure~\ref{fig:case_study} illustrates how evolved experience improves reasoning behavior. For a problem requiring the last two digits to satisfy both a digit-sum constraint and the divisibility-by-4 rule, the baseline model fails to systematically enumerate all candidates whose digits sum to 13, introduces several invalid candidates, and ultimately produces the incorrect answer. After incorporating the experience, the model adopts a more structured two-stage reasoning procedure. It first exhaustively enumerates all ordered digit pairs, and then verifies whether each corresponding two-digit number is divisible by $4$. This procedure yields the correct result $7\times6=42$. Importantly, the injected experience contains neither the answer $42$ nor any candidates specific to this problem. Instead, it provides transferable reasoning principles, such as decomposing the problem into smaller steps. Therefore, the improvement does not result from answer memorization, but from the adoption of a more reliable reasoning procedure.

\section{Related Work}

\subsection{Prompt-Based Experience Utilization}
LLMs exhibit strong in-context learning capabilities, allowing them to adapt their behaviors by leveraging external information provided in prompts. Methods such as chain-of-thought prompting and self-consistency demonstrate that demonstrations and reasoning trajectories can serve as explicit experience to guide model reasoning~\cite{yao2023tree,wang2022self,besta2024graph}. Retrieval-augmented generation further extends this paradigm by incorporating additional knowledge from external databases to enhance model capabilities~\cite{NEURIPS2020_6b493230,jin2025searchr1trainingllmsreason}. Recent studies on agents introduce long-term memory mechanisms that store historical observations, actions, and reflection results, enabling long-horizon interactions~\cite{park2023generative,zhao2024expel,shinn2023reflexion}. However, such experience remains external to model parameters, and its effectiveness depends on retrieval quality and context capacity. Once the relevant information is removed from the context, the acquired knowledge cannot be directly preserved as intrinsic model capabilities~\cite{liu-etal-2024-lost}.

\subsection{Reinforcement Learning}

Reinforcement learning~(RL) improves the capabilities of large language models (LLMs) by optimizing model policies through environmental feedback. Early approaches, such as reinforcement learning from human feedback (RLHF), align model behaviors with human preferences by learning reward signals and applying policy optimization algorithms~\cite{ouyang2022training,stiennon2022learningsummarizehumanfeedback,ziegler2020finetuninglanguagemodelshuman}. With the emergence of automatically verifiable tasks, reinforcement learning has been widely applied to reasoning tasks, where rewards can be obtained through verifiers, or rule-based evaluators~\cite{luong2024reftreasoningreinforcedfinetuning,shao2024deepseekmathpushinglimitsmathematical,le2022coderlmasteringcodegeneration}. Large-scale reward-driven training enables models to autonomously develop complex behaviors, including longer reasoning processes,  backtracking, and strategy adjustment. These advances have gradually transformed reinforcement learning from a preference alignment technique into an important post-training paradigm for enhancing reasoning and exploration capabilities~\cite{kimiteam2025kimik15scalingreinforcement,jin2025searchr1trainingllmsreason,wang2025ragenunderstandingselfevolutionllm}. However, reinforcement learning heavily relies on the strength of the initial model, while experience is primarily absorbed through scalar reward signals. When the model capability is limited, sparse rewards and inefficient exploration can lead to suboptimal performance.

\subsection{On-Policy Self-Distillation}

Knowledge distillation transfers knowledge from a teacher model to a student model, enabling effective knowledge transfer between models~\cite{hinton2015distillingknowledgeneuralnetwork}. On-policy distillation further improves this process by sampling trajectories from the current policy, ensuring that the supervision signal matches the behavior distribution of the student model itself~\cite{agarwal2024onpolicydistillationlanguagemodels,gu2026minillmonpolicydistillationlarge}. In self-evolving scenarios, existing methods typically construct teacher models using privileged information and directly distill complete answers, reasoning trajectories, or teacher response distributions~\cite{penaloza2026privilegedinformationdistillationlanguage,zhao2026selfdistilledreasoneronpolicyselfdistillation}. However, such trajectory-level supervision signals often contain substantial problem-specific details, causing student models to imitate particular solution processes rather than learn generalizable strategies shared across different examples~\cite{shi2026trajectoryimitationstrategyguidedpolicy}. Compared with these methods, we introduce a continuously evolving experience representation that extracts, integrates, and filters transferable knowledge, thereby enabling more effective experience internalization.

\section{Conclusion and Limitation} 
In this work, we introduce SPEE, a unified post-training framework for transforming transient interaction experience into persistent model capabilities. SPEE closes the loop between explicit experience utilization and implicit policy optimization by extracting and refining transferable experience, internalizing it through self-distillation, and further improving the policy through reinforcement learning. Experiments across five mathematical reasoning benchmarks and three model scales demonstrate consistent gains over test-time and training-time self-evolution baselines. 
Future work may extend this experience-evolution paradigm to a broader range of task domains and application scenarios with sparser, less reliable, or more complex environmental feedback. The framework supports closed-loop iteration, future work may further explore multiple rounds of methods.

\bibliography{aaai2027}

@article{achiam2023gpt,
  title={Gpt-4 technical report},
  author={Achiam, Josh and Adler, Steven and Agarwal, Sandhini and Ahmad, Lama and Akkaya, Ilge and Aleman, Florencia Leoni and Almeida, Diogo and Altenschmidt, Janko and Altman, Sam and Anadkat, Shyamal and others},
  journal={arXiv preprint arXiv:2303.08774},
  year={2023}
}

@article{wei2022chain,
  title={Chain-of-thought prompting elicits reasoning in large language models},
  author={Wei, Jason and Wang, Xuezhi and Schuurmans, Dale and Bosma, Maarten and Xia, Fei and Chi, Ed and Le, Quoc V and Zhou, Denny and others},
  journal={Advances in neural information processing systems},
  volume={35},
  pages={24824--24837},
  year={2022}
}

@article{yao2022react,
  title={React: Synergizing reasoning and acting in language models},
  author={Yao, Shunyu and Zhao, Jeffrey and Yu, Dian and Du, Nan and Shafran, Izhak and Narasimhan, Karthik and Cao, Yuan},
  journal={arXiv preprint arXiv:2210.03629},
  year={2022}
}

@article{ouyang2022training,
  title={Training language models to follow instructions with human feedback},
  author={Ouyang, Long and Wu, Jeffrey and Jiang, Xu and Almeida, Diogo and Wainwright, Carroll and Mishkin, Pamela and Zhang, Chong and Agarwal, Sandhini and Slama, Katarina and Ray, Alex and others},
  journal={Advances in neural information processing systems},
  volume={35},
  pages={27730--27744},
  year={2022}
}

@article{wang2023voyager,
  title={Voyager: An open-ended embodied agent with large language models},
  author={Wang, Guanzhi and Xie, Yuqi and Jiang, Yunfan and Mandlekar, Ajay and Xiao, Chaowei and Zhu, Yuke and Fan, Linxi and Anandkumar, Anima},
  journal={arXiv preprint arXiv:2305.16291},
  year={2023}
}

@article{kaplan2020scaling,
  title={Scaling laws for neural language models},
  author={Kaplan, Jared and McCandlish, Sam and Henighan, Tom and Brown, Tom B and Chess, Benjamin and Child, Rewon and Gray, Scott and Radford, Alec and Wu, Jeffrey and Amodei, Dario},
  journal={arXiv preprint arXiv:2001.08361},
  year={2020}
}

@misc{tao2survey,
  title={A survey on self-evolution of large language models. arXiv. 240414387 (2024)},
  year={2024},
  author={Tao, Z and Lin, TE and Chen, X and Li, H and Wu, Y and Li, Y and Jin, Z and Huang, F and Tao, D and Zhou, J}
}

@article{mohamed2025context,
  title={In-context learning in large language models (LLMs): Mechanisms, capabilities, and implications for advanced knowledge representation and reasoning},
  author={Mohamed, Azza and El Rashid, Mohamed and Shaalan, Khaled},
  journal={IEEE Access},
  volume={13},
  pages={95574--95593},
  year={2025},
  publisher={IEEE}
}

@article{debnath2025comprehensive,
  title={A Comprehensive Survey of Prompt Engineering Techniques in Large Language Models},
  author={Debnath, Tonmoy and Siddiky, Nurul Absar and Rahman, Muhammad Enayetur and Das, Prosenjit and Guha, Antu Kumar},
  journal={Authorea Preprints},
  year={2025},
  publisher={Authorea}
}

@article{guo2025deepseek,
  title={Deepseek-r1: Incentivizing reasoning capability in llms via reinforcement learning},
  author={Guo, Daya and Yang, Dejian and Zhang, Haowei and Song, Junxiao and Wang, Peiyi and Zhu, Qihao and Xu, Runxin and Zhang, Ruoyu and Ma, Shirong and Bi, Xiao and others},
  journal={arXiv preprint arXiv:2501.12948},
  year={2025}
}

@article{yu2026dapo,
  title={Dapo: An open-source llm reinforcement learning system at scale},
  author={Yu, Qiying and Zhang, Zheng and Zhu, Ruofei and Yuan, Yufeng and Zuo, Xiaochen and Yue, Yu and Dai, Weinan and Fan, Tiantian and Liu, Gaohong and Liu, Lingjun and others},
  journal={Advances in Neural Information Processing Systems},
  volume={38},
  pages={113222--113244},
  year={2026}
}

@INPROCEEDINGS{wang2026autonomous,
  author={Wang, Feiyang and Ma, Yumeng and Guan, Tian and Wang, Yutong and Chen, Jinyu},
  booktitle={2026 International Conference on Embedded Systems, Mobile Communication and Computing (EMC²)}, 
  title={Autonomous Learning through Self-Driven Exploration and Knowledge Structuring for Open-World Intelligent Agents}, 
  year={2026},
  volume={},
  number={},
  pages={301-306},
  doi={10.1109/EMC68537.2026.11441790}}

@misc{zhao2026selfdistilledreasoneronpolicyselfdistillation,
      title={Self-Distilled Reasoner: On-Policy Self-Distillation for Large Language Models}, 
      author={Siyan Zhao and Zhihui Xie and Mengchen Liu and Jing Huang and Guan Pang and Feiyu Chen and Aditya Grover},
      year={2026},
      eprint={2601.18734},
      archivePrefix={arXiv},
      primaryClass={cs.LG},
      url={https://arxiv.org/abs/2601.18734}, 
}

@misc{wang2023selfconsistencyimproveschainthought,
      title={Self-Consistency Improves Chain of Thought Reasoning in Language Models}, 
      author={Xuezhi Wang and Jason Wei and Dale Schuurmans and Quoc Le and Ed Chi and Sharan Narang and Aakanksha Chowdhery and Denny Zhou},
      year={2023},
      eprint={2203.11171},
      archivePrefix={arXiv},
      primaryClass={cs.CL},
      url={https://arxiv.org/abs/2203.11171}, 
}

@misc{shao2024deepseekmathpushinglimitsmathematical,
      title={DeepSeekMath: Pushing the Limits of Mathematical Reasoning in Open Language Models}, 
      author={Zhihong Shao and Peiyi Wang and Qihao Zhu and Runxin Xu and Junxiao Song and Xiao Bi and Haowei Zhang and Mingchuan Zhang and Y. K. Li and Y. Wu and Daya Guo},
      year={2024},
      eprint={2402.03300},
      archivePrefix={arXiv},
      primaryClass={cs.CL},
      url={https://arxiv.org/abs/2402.03300}, 
}

@article{liu-etal-2024-lost,
    title = "Lost in the Middle: How Language Models Use Long Contexts",
    author = "Liu, Nelson F.  and
      Lin, Kevin  and
      Hewitt, John  and
      Paranjape, Ashwin  and
      Bevilacqua, Michele  and
      Petroni, Fabio  and
      Liang, Percy",
    journal = "Transactions of the Association for Computational Linguistics",
    volume = "12",
    year = "2024",
    address = "Cambridge, MA",
    publisher = "MIT Press",
    url = "https://aclanthology.org/2024.tacl-1.9/",
    doi = "10.1162/tacl_a_00638",
    pages = "157--173"
}

@inproceedings{mueller-etal-2024-context,
    title = "In-context Learning Generalizes, But Not Always Robustly: The Case of Syntax",
    author = "Mueller, Aaron  and
      Webson, Albert  and
      Petty, Jackson  and
      Linzen, Tal",
    editor = "Duh, Kevin  and
      Gomez, Helena  and
      Bethard, Steven",
    booktitle = "Proceedings of the 2024 Conference of the North American Chapter of the Association for Computational Linguistics: Human Language Technologies (Volume 1: Long Papers)",
    month = jun,
    year = "2024",
    address = "Mexico City, Mexico",
    publisher = "Association for Computational Linguistics",
    url = "https://aclanthology.org/2024.naacl-long.267/",
    doi = "10.18653/v1/2024.naacl-long.267",
    pages = "4761--4779"
}

@article{yao2023tree,
  title={Tree of thoughts: Deliberate problem solving with large language models},
  author={Yao, Shunyu and Yu, Dian and Zhao, Jeffrey and Shafran, Izhak and Griffiths, Tom and Cao, Yuan and Narasimhan, Karthik},
  journal={Advances in neural information processing systems},
  volume={36},
  pages={11809--11822},
  year={2023}
}

@article{wang2022self,
  title={Self-consistency improves chain of thought reasoning in language models},
  author={Wang, Xuezhi and Wei, Jason and Schuurmans, Dale and Le, Quoc and Chi, Ed and Narang, Sharan and Chowdhery, Aakanksha and Zhou, Denny},
  journal={arXiv preprint arXiv:2203.11171},
  year={2022}
}

@inproceedings{besta2024graph,
  title={Graph of thoughts: Solving elaborate problems with large language models},
  author={Besta, Maciej and Blach, Nils and Kubicek, Ales and Gerstenberger, Robert and Podstawski, Michal and Gianinazzi, Lukas and Gajda, Joanna and Lehmann, Tomasz and Niewiadomski, Hubert and Nyczyk, Piotr and others},
  booktitle={Proceedings of the AAAI conference on artificial intelligence},
  volume={38},
  pages={17682--17690},
  year={2024}
}

@inproceedings{NEURIPS2020_6b493230,
 author = {Lewis, Patrick and Perez, Ethan and Piktus, Aleksandra and Petroni, Fabio and Karpukhin, Vladimir and Goyal, Naman and K\"{u}ttler, Heinrich and Lewis, Mike and Yih, Wen-tau and Rockt\"{a}schel, Tim and Riedel, Sebastian and Kiela, Douwe},
 booktitle = {Advances in Neural Information Processing Systems},
 editor = {H. Larochelle and M. Ranzato and R. Hadsell and M.F. Balcan and H. Lin},
 pages = {9459--9474},
 publisher = {Curran Associates, Inc.},
 title = {Retrieval-Augmented Generation for Knowledge-Intensive NLP Tasks},
 url = {https://proceedings.neurips.cc/paper_files/paper/2020/file/6b493230205f780e1bc26945df7481e5-Paper.pdf},
 volume = {33},
 year = {2020}
}

@misc{jin2025searchr1trainingllmsreason,
      title={Search-R1: Training LLMs to Reason and Leverage Search Engines with Reinforcement Learning}, 
      author={Bowen Jin and Hansi Zeng and Zhenrui Yue and Jinsung Yoon and Sercan Arik and Dong Wang and Hamed Zamani and Jiawei Han},
      year={2025},
      eprint={2503.09516},
      archivePrefix={arXiv},
      primaryClass={cs.CL},
      url={https://arxiv.org/abs/2503.09516}, 
}

@inproceedings{park2023generative,
  title={Generative agents: Interactive simulacra of human behavior},
  author={Park, Joon Sung and O'Brien, Joseph and Cai, Carrie Jun and Morris, Meredith Ringel and Liang, Percy and Bernstein, Michael S},
  booktitle={Proceedings of the 36th annual acm symposium on user interface software and technology},
  pages={1--22},
  year={2023}
}

@inproceedings{zhao2024expel,
  title={Expel: Llm agents are experiential learners},
  author={Zhao, Andrew and Huang, Daniel and Xu, Quentin and Lin, Matthieu and Liu, Yong-Jin and Huang, Gao},
  booktitle={Proceedings of the AAAI Conference on Artificial Intelligence},
  volume={38},
  pages={19632--19642},
  year={2024}
}

@article{shinn2023reflexion,
  title={Reflexion: Language agents with verbal reinforcement learning},
  author={Shinn, Noah and Cassano, Federico and Gopinath, Ashwin and Narasimhan, Karthik and Yao, Shunyu},
  journal={Advances in neural information processing systems},
  volume={36},
  pages={8634--8652},
  year={2023}
}

@misc{stiennon2022learningsummarizehumanfeedback,
      title={Learning to summarize from human feedback}, 
      author={Nisan Stiennon and Long Ouyang and Jeff Wu and Daniel M. Ziegler and Ryan Lowe and Chelsea Voss and Alec Radford and Dario Amodei and Paul Christiano},
      year={2022},
      eprint={2009.01325},
      archivePrefix={arXiv},
      primaryClass={cs.CL},
      url={https://arxiv.org/abs/2009.01325}, 
}

@misc{ziegler2020finetuninglanguagemodelshuman,
      title={Fine-Tuning Language Models from Human Preferences}, 
      author={Daniel M. Ziegler and Nisan Stiennon and Jeffrey Wu and Tom B. Brown and Alec Radford and Dario Amodei and Paul Christiano and Geoffrey Irving},
      year={2020},
      eprint={1909.08593},
      archivePrefix={arXiv},
      primaryClass={cs.CL},
      url={https://arxiv.org/abs/1909.08593}, 
}

@misc{luong2024reftreasoningreinforcedfinetuning,
      title={ReFT: Reasoning with Reinforced Fine-Tuning}, 
      author={Trung Quoc Luong and Xinbo Zhang and Zhanming Jie and Peng Sun and Xiaoran Jin and Hang Li},
      year={2024},
      eprint={2401.08967},
      archivePrefix={arXiv},
      primaryClass={cs.CL},
      url={https://arxiv.org/abs/2401.08967}, 
}

@misc{le2022coderlmasteringcodegeneration,
      title={CodeRL: Mastering Code Generation through Pretrained Models and Deep Reinforcement Learning}, 
      author={Hung Le and Yue Wang and Akhilesh Deepak Gotmare and Silvio Savarese and Steven C. H. Hoi},
      year={2022},
      eprint={2207.01780},
      archivePrefix={arXiv},
      primaryClass={cs.LG},
      url={https://arxiv.org/abs/2207.01780}, 
}

@misc{kimiteam2025kimik15scalingreinforcement,
      title={Kimi k1.5: Scaling Reinforcement Learning with LLMs}, 
      author={Kimi Team and Angang Du and Bofei Gao and Bowei Xing and Changjiu Jiang and Cheng Chen and Cheng Li and Chenjun Xiao and Chenzhuang Du and Chonghua Liao and Chuning Tang and Congcong Wang and Dehao Zhang and Enming Yuan and Enzhe Lu and Fengxiang Tang and Flood Sung and Guangda Wei and Guokun Lai and Haiqing Guo and Han Zhu and Hao Ding and Hao Hu and Hao Yang and Hao Zhang and Haotian Yao and Haotian Zhao and Haoyu Lu and Haoze Li and Haozhen Yu and Hongcheng Gao and Huabin Zheng and Huan Yuan and Jia Chen and Jianhang Guo and Jianlin Su and Jianzhou Wang and Jie Zhao and Jin Zhang and Jingyuan Liu and Junjie Yan and Junyan Wu and Lidong Shi and Ling Ye and Longhui Yu and Mengnan Dong and Neo Zhang and Ningchen Ma and Qiwei Pan and Qucheng Gong and Shaowei Liu and Shengling Ma and Shupeng Wei and Sihan Cao and Siying Huang and Tao Jiang and Weihao Gao and Weimin Xiong and Weiran He and Weixiao Huang and Weixin Xu and Wenhao Wu and Wenyang He and Xianghui Wei and Xianqing Jia and Xingzhe Wu and Xinran Xu and Xinxing Zu and Xinyu Zhou and Xuehai Pan and Y. Charles and Yang Li and Yangyang Hu and Yangyang Liu and Yanru Chen and Yejie Wang and Yibo Liu and Yidao Qin and Yifeng Liu and Ying Yang and Yiping Bao and Yulun Du and Yuxin Wu and Yuzhi Wang and Zaida Zhou and Zhaoji Wang and Zhaowei Li and Zhen Zhu and Zheng Zhang and Zhexu Wang and Zhilin Yang and Zhiqi Huang and Zihao Huang and Ziyao Xu and Zonghan Yang and Zongyu Lin},
      year={2025},
      eprint={2501.12599},
      archivePrefix={arXiv},
      primaryClass={cs.AI},
      url={https://arxiv.org/abs/2501.12599}, 
}

@misc{wang2025ragenunderstandingselfevolutionllm,
      title={RAGEN: Understanding Self-Evolution in LLM Agents via Multi-Turn Reinforcement Learning}, 
      author={Zihan Wang and Kangrui Wang and Qineng Wang and Pingyue Zhang and Linjie Li and Zhengyuan Yang and Xing Jin and Kefan Yu and Minh Nhat Nguyen and Licheng Liu and Eli Gottlieb and Yiping Lu and Kyunghyun Cho and Jiajun Wu and Li Fei-Fei and Lijuan Wang and Yejin Choi and Manling Li},
      year={2025},
      eprint={2504.20073},
      archivePrefix={arXiv},
      primaryClass={cs.LG},
      url={https://arxiv.org/abs/2504.20073}, 
}

@misc{hinton2015distillingknowledgeneuralnetwork,
      title={Distilling the Knowledge in a Neural Network}, 
      author={Geoffrey Hinton and Oriol Vinyals and Jeff Dean},
      year={2015},
      eprint={1503.02531},
      archivePrefix={arXiv},
      primaryClass={stat.ML},
      url={https://arxiv.org/abs/1503.02531}, 
}

@misc{agarwal2024onpolicydistillationlanguagemodels,
      title={On-Policy Distillation of Language Models: Learning from Self-Generated Mistakes}, 
      author={Rishabh Agarwal and Nino Vieillard and Yongchao Zhou and Piotr Stanczyk and Sabela Ramos and Matthieu Geist and Olivier Bachem},
      year={2024},
      eprint={2306.13649},
      archivePrefix={arXiv},
      primaryClass={cs.LG},
      url={https://arxiv.org/abs/2306.13649}, 
}

@misc{gu2026minillmonpolicydistillationlarge,
      title={MiniLLM: On-Policy Distillation of Large Language Models}, 
      author={Yuxian Gu and Li Dong and Furu Wei and Minlie Huang},
      year={2026},
      eprint={2306.08543},
      archivePrefix={arXiv},
      primaryClass={cs.CL},
      url={https://arxiv.org/abs/2306.08543}, 
}

@misc{penaloza2026privilegedinformationdistillationlanguage,
      title={Privileged Information Distillation for Language Models}, 
      author={Emiliano Penaloza and Dheeraj Vattikonda and Nicolas Gontier and Alexandre Lacoste and Laurent Charlin and Massimo Caccia},
      year={2026},
      eprint={2602.04942},
      archivePrefix={arXiv},
      primaryClass={cs.LG},
      url={https://arxiv.org/abs/2602.04942}, 
}

@misc{shi2026trajectoryimitationstrategyguidedpolicy,
      title={Beyond Trajectory Imitation: Strategy-Guided Policy Optimization for LLM Reasoning}, 
      author={Tianyuan Shi and Canbin Huang and Bei Li and Xin Chen and Xiaojun Quan and Jingang Wang and Qifan Wang},
      year={2026},
      eprint={2606.24064},
      archivePrefix={arXiv},
      primaryClass={cs.AI},
      url={https://arxiv.org/abs/2606.24064}, 
}

@misc{cobbe2021trainingverifierssolvemath,
      title={Training Verifiers to Solve Math Word Problems}, 
      author={Karl Cobbe and Vineet Kosaraju and Mohammad Bavarian and Mark Chen and Heewoo Jun and Lukasz Kaiser and Matthias Plappert and Jerry Tworek and Jacob Hilton and Reiichiro Nakano and Christopher Hesse and John Schulman},
      year={2021},
      eprint={2110.14168},
      archivePrefix={arXiv},
      primaryClass={cs.LG},
      url={https://arxiv.org/abs/2110.14168}, 
}

@misc{aime24,
      title={American Invitational Mathematics Examination (AIME) 2024}, 
      author={Zhang, Yifan and Math-AI, Team},
      year={2024},
}

@misc{aime25,
  title     = {American Invitational Mathematics Examination (AIME) 2025},
  author    = {Zhang, Yifan and Math-AI Team},
  year      = {2025},
  publisher = {HuggingFace},
  url       = {https://huggingface.co/datasets/math-ai/aime25}
}

@article{lightman2023lets,
      title={Let's Verify Step by Step}, 
      author={Lightman, Hunter and Kosaraju, Vineet and Burda, Yura and Edwards, Harri and Baker, Bowen and Lee, Teddy and Leike, Jan and Schulman, John and Sutskever, Ilya and Cobbe, Karl},
      journal={arXiv preprint arXiv:2305.20050},
      year={2023}
}

@article{lewkowycz2022solving,
  title={Solving quantitative reasoning problems with language models},
  author={Lewkowycz, Aitor and Andreassen, Anders and Dohan, David and Dyer, Ethan and Michalewski, Henryk and Ramasesh, Vinay and Slone, Ambrose and Anil, Cem and Schlag, Imanol and Gutman-Solo, Theo and others},
  journal={Advances in neural information processing systems},
  volume={35},
  pages={3843--3857},
  year={2022}
}

@inproceedings{hubotter2026reinforcement,
title={Reinforcement Learning via Self-Distillation},
author={Jonas H{\"u}botter and Frederike L{\"u}beck and Lejs Deen Behric and Anton Baumann and Marco Bagatella and Daniel Marta and Ido Hakimi and Idan Shenfeld and Thomas Kleine Buening and Carlos Guestrin and Andreas Krause},
booktitle={Forty-third International Conference on Machine Learning},
year={2026},
url={https://openreview.net/forum?id=QkfkxyRizZ}
}

\end{document}